\documentclass[conference]{IEEEtran}
\IEEEoverridecommandlockouts

\usepackage{amsmath,amssymb,amsfonts}

\usepackage{booktabs}
\usepackage[table,xcdraw,dvipsnames]{xcolor}
\usepackage[switch]{lineno}

\usepackage{soul}
\usepackage[utf8]{inputenc}
\usepackage[small]{caption}
\usepackage{subcaption}
\usepackage{multirow}
\usepackage{array}
\usepackage{float}
\usepackage[table]{xcolor}
\definecolor{mybluecolor}{HTML}{80C4E9}
\usepackage{makecell}
\usepackage{times}
\usepackage{tabularx}

\usepackage{times}  
\usepackage{helvet}  
\usepackage{courier}  
\usepackage[hyphens]{url}  
\usepackage{graphicx} 
\usepackage{caption} 

\usepackage{algorithm}
\usepackage{algorithmic}

\usepackage{amsthm}
\usepackage{amssymb}
\usepackage{comment} 
\usepackage{amsmath} 
\usepackage{booktabs}
\usepackage{colortbl}
\usepackage{xcolor} 
\usepackage{adjustbox}

\usepackage[most, breakable]{tcolorbox}
\usepackage{enumitem}
\usepackage{amsthm}

\newtheorem{example}{Example}

\usepackage[backend=bibtex]{biblatex}
\def\BibTeX{{\rm B\kern-.05em{\sc i\kern-.025em b}\kern-.08em
    T\kern-.1667em\lower.7ex\hbox{E}\kern-.125emX}}
\begin{document}

\title{\huge Aligning Knowledge Graphs and Language Models for Factual Accuracy
\thanks{The opinions expressed in this paper are personal and should not be attributed to Banca d'Italia.}
\thanks{\textsuperscript{*}The work was done outside Amazon.} 
}

\author{
\IEEEauthorblockN{
Nur A Zarin Nishat\IEEEauthorrefmark{1},
Andrea Coletta\IEEEauthorrefmark{2},
Luigi Bellomarini\IEEEauthorrefmark{2},\\
Kossi Amouzouvi\IEEEauthorrefmark{3}
Jens Lehmann\IEEEauthorrefmark{3}\IEEEauthorrefmark{4},
Sahar Vahdati\IEEEauthorrefmark{1}
}
\\

\IEEEauthorblockA{\IEEEauthorrefmark{1}TIB – Leibniz Information Centre for Science and Technology, Hannover, Germany  \\ \{nur.nishat, sahar.vahdati\}@tib.eu}
\IEEEauthorblockA{\IEEEauthorrefmark{3}ScaDS.AI Dresden/Leipzig, Technische Universität Dresden, 01069 Dresden, Germany \\
\{jens.lehmann, kossi.amouzouvi\}@tu-dresden.de}

\IEEEauthorblockA{\IEEEauthorrefmark{2}Banca d’Italia, Rome, Italy \\
\{andrea.coletta, luigi.bellomarini\}@bancaditalia.it}

\IEEEauthorblockA{\IEEEauthorrefmark{4}Amazon, Germany\textsuperscript{*}} 
}

\maketitle

\begin{abstract}
Large language models like GPT-4, Gemini, and Claude have transformed natural language processing (NLP) tasks such as question answering, dialogue generation, summarization, and so forth; yet their susceptibility to hallucination stands as one of the major challenges. Among numerous approaches to overcome this challenge, integration of Knowledge Graphs (KGs) into language models has emerged as a promising solution as it provides structured, reliable, domain-specific, and up-to-date external information to the language models. In this paper, we introduce ALIGNed-LLM, a simple yet effective approach to improve language models' factuality via a lean strategy to infuse KGs into the latent space of language models inspired by LLaVA where visual and textual information is infused. We use embeddings from a pre-trained Knowledge Graph Embedding (KGE) model, such as TransE, and a trainable projection layer to align entity and text embeddings. This alignment enables the language model to distinguish between similar entities improving factual grounding and reducing hallucination. We tested our approach on three popular questions-answering benchmark datasets alongside language models of varying sizes, showing significant improvement. Furthermore, we applied our approach to a real-world financial use case from a large central bank in Europe, which demands high accuracy and precision, demonstrating a substantial improvement of the LLM answers. 
\end{abstract}


\section{Introduction}\label{sec:intro}

The emergence of Large Language Models (LLMs) such as GPT-4~\cite{openai2024gpt}, Gemini~\cite{anil2023gemini}, Llama~\cite{touvron2023llama}, and Claude~\cite{anthropic2023claude}, is producing a performance revolution across natural language processing tasks, such as question answering, dialogue generation, summarization, and many more. LLMs have also confirmed the universality of human language, showing their potential to build domain-specific assistants, trained to follow natural language instructions and accomplish various tasks end to end~\cite{liu2024visual}.
Furthermore, LLMs are sparking discussions about the Artificial Intelligence foundations~\cite{bubeck2023sparks}, thanks to their ability to solve reasoning- and generalization-based tasks in many fields, spanning mathematics, coding, vision, medicine, law, psychology, and more~\cite{bubeck2023sparks}.  
The new stream of approaches is positioning LLMs as a pillar for neural-symbolic techniques~\cite{DBLP:conf/nesy/2024-1}, particularly suitable in realms such as finance~\cite{DBLP:conf/rulemlrr/BaldazziBBSV24}, characterized by consolidated and strict regulations or domain rules.
\textit{Knowledge Representation and Reasoning} techniques and semantic data-structures such as \textit{Knowledge Graphs} (KG) can be adopted to reduce the well-known factuality lacks of LLMs with the flexibility of language-based thinking.

Indeed, factuality problems---due, for example, to stale training data~\cite{peng2023check}, propensity to hallucinations~\cite{ji2023survey}, lack of domain expertise~\cite{zheng2024fine}---are still a key limitation of LLMs.

\begin{figure*}[htbp]
\centerline{\includegraphics[scale=.5]{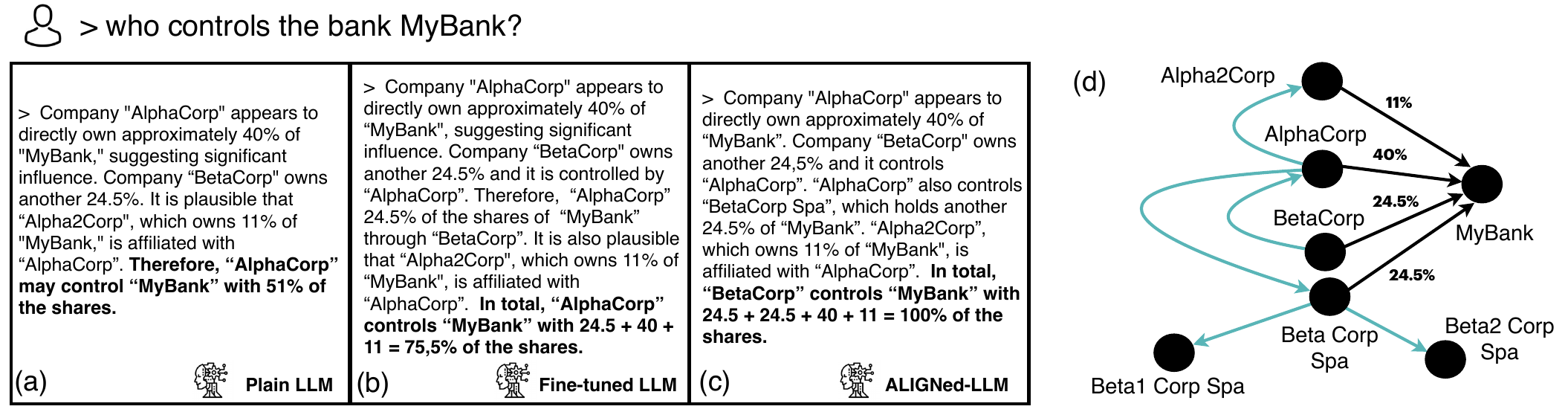}}
    \caption{A question answering setting over the company control scenario from the central bank, addressed with: (a) a plain LLM; (b) an LLM fine-tuned with the ownership KG of the central bank; (c) an ALIGNed-LLM using our approach. An anonymized fragment of the ownership KG is shown (d); ownership edges, weighted with the shares, are in black, while direct control edges are in blue.}
    \label{fig:running_example}
\end{figure*}

\smallskip
Consider, for example, the following case from a large central bank in Europe, where the goal is to determine who ultimately controls the bank \texttt{MyBank}, according to the following recursive definition~\cite{DBLP:conf/icde/GulinoCGSB21}. 
\textit{A bank, or, more in general, a company X is considered to control Company Y if X either directly owns more than 50\% of Y or indirectly does so by controlling a group of companies that collectively own more than 50\% of Y. If X is not controlled by any other company, it is named as the ``ultimate controller'' of Y.} 

\begin{example}
\textit{Consider Figure~\ref{fig:running_example}(a). 
We want to use a plain LLM, not exposed to the KG of the central bank, to determine the ultimate controller of \texttt{MyBank}. While the model correctly guesses that \texttt{Alpha2Corp} is a subsidiary of \texttt{AlphaCorp}, it incorrectly concludes that \texttt{AlphaCorp} controls \texttt{MyBank}, as it seems to be unaware (stale training data) that \texttt{BetaCorp} in turn controls \texttt{AlphaCorp}. 
} \hfill$\blacksquare$
\label{ex:running_example}
\end{example}
Although hallucinations can be considered inherent characteristics of LLM architectures~\cite{banerjee2024llms}, approaches have emerged that try to mitigate this phenomenon, unacceptable in critical domains. 
The integration of external structured knowledge sources such as KGs into LLM architectures is a promising direction. Knowledge Graphs are structured repositories of real-world knowledge, representing entities and the relationships between them, and providing factual information in an organized format~\cite{wang2017knowledge}. 
By integrating with KGs, Large Language Models gain access to a comprehensive and up-to-date source of domain-specific knowledge, enabling them to ground their reasoning processes in real-world contexts, such as business environments, and significantly enhance their accuracy and relevance~\cite{wu2024thinking}. 

The integration between KGs and LLMs is studied along two directions: (i) \textit{fine-tuning} on domain-specific KGs, so as to align the model output more closely with the factual knowledge~\cite{DBLP:conf/nips/BrownMRSKDNSSAA20}; (ii) augmenting the LLM with \textit{Retrieval-Augmented Generation} (RAG or GraphRAG) techniques where subgraphs or triples are retrieved from the KG and included in the prompt, providing further context to the LLM~\cite{mallen2022not,baek2023knowledge} like in the case of broad general purpose KGs, such as Wikidata and Wikipedia.
Despite the promise of fine-tuning and RAG approaches, linearizing KGs into text format often leads to a loss of structure information and is not able to fully utilize the relational context~\cite{guo2024knowledgenavigator}, sometimes resulting in irrelevant retrieved information to the question and ungrounded generated answer with respect to the retrieved knowledge~\cite{baek2023knowledge}. Moreover, prompt-based methods can encounter information conflicts and thus can be ignored by the pre-existing biases of the LLMs~\cite{wang2023resolving}. 
Finally, the computational cost of large-scale LLMs like GPT-4 can exacerbate these challenges. 

\begin{example}
\label{ex:running_example_fine-tuning}
\textit{In Figure~\ref{fig:running_example}(b), an LLM fine-tuned with KG data still incorrectly concludes that \texttt{AlphaCorp} controls \texttt{MyBank}, since the model mistakes (hallucination) \texttt{BetaCorp}, which controls \texttt{AlphaCorp}, for its subsidiary \texttt{Beta Corp Spa}. After the fine-tuning process, the model fails to disambiguate these two entities, \texttt{Beta Corp Spa} and \texttt{BetaCorp}, likely due to their similar names and relationships. 
} \hfill$\blacksquare$ 
\end{example}

\smallskip    
A simple yet effective way of bridging two different modalities to represent knowledge was demonstrated by LLaVA~\cite{liu2024visual}, where visual data and textual instruction are combined to enhance natural language question answering. 
LLaVA leverages the parameter-efficient fine-tuning (PEFT) and alignment of visual and textual embedding spaces via a trainable projection layer and achieves significant improvements in understanding and reasoning with multi-modal inputs. 
In this paper, we borrow and extend this concept to address the challenges of incorporating KGs with language models.
Our approach aligns structured knowledge from KGs with the text embedding space of LLMs. In particular, we leverage pre-trained \textit{Knowledge Graph Embeddings} (KGEs), i.e., encodings of the KG entities as low-dimensional vector spaces capturing their semantic and structural properties~\cite{DBLP:journals/tnn/JiPCMY22}, and align them with text embeddings through a lightweight projection layer. 
Our infusion technique improves the ability of the LLM to resolve ambiguities and differentiate between similar entities, thanks to a rich semantic context, such as distinguishing \texttt{BetaCorp} and \texttt{Beta Corp Spa} in Figure~\ref{fig:running_example}(c).

\smallskip\noindent\textbf{Contribution}. We summarize our main contributions as:

\begin{itemize}
    \item The \textit{architecture of the \textup{ALIGNed-LLM} system}, featuring the infusion of KG knowledge into the latent space of LLMs thanks to KGEs. 
    \item A lightweight \textit{projection layer} that aligns the embedding of the reference entities with the embedding space of the Language model.
    \item A \textit{quantitative evaluation} of the performance of our framework across a variety of QA datasets---both public and enterprise KGs, such as the ownership KG of the central bank---showcasing its ability to improve factual accuracy and mitigate entity-related errors.
\end{itemize}

\section{Related Work}
\label{sec:relwork}
The infusion of knowledge graphs with large language models has emerged as a promising approach to reducing hallucinations in natural language processing tasks. 
Prior work has demonstrated the benefits of this infusion, enhancing the AI system’s factual accuracy and reasoning capabilities. Various methods have been proposed to incorporate knowledge graphs into large language models.

Pan et al.~\cite{pan2024unifying} provide a comprehensive roadmap, including directions for future research, for unifying large language models and knowledge graphs, discussing various theoretical frameworks that leverage the strengths of both to enhance AI interpretability and inference capabilities.

A novel approach to integrating KGs into LLMs is introduced by KG-Adapter~\cite{tian2024kg}, which encodes KG structure information by incorporating both node-centric and relation-centric data, as well as triples, using GNN and MLP. 
It ensures computational efficiency through a PEFT-based fine-tuning method designed for decoder-only LLMs.

ChatKBQA~\cite{luo2023chatkbqa} introduces a Generate-then-Retrieve framework, where the logical form for natural language question answering is first generated using fine-tuned LLMs.
Entities and relations are then retrieved and substituted through an unsupervised retrieval method.

REANO~\cite{fang2024reano} proposes to enhance the retrieval-augmented reader model with a knowledge graph-generating module. It generates a KG from the given passages and uses a GNN to identify the top-k triples most relevant to the questions. 
Together with the input text and the question, these triples are then encoded and provided to a decoder to predict the answer.

ELPF~\cite{jiang2024efficient} framework introduces efficient infusion of KGs. With a small set of labeled samples and a large-scale corpus domain, a domain-specific KG is constructed using a fine-tuned LLM.
Also, they employ a three-stage KG-LLM alignment strategy, where in the final stage, KG is provided as an automated evaluator for knowledge correctness and better alignment. This work focuses on biomedical question-answering datasets.

KnowLA~\cite{luo2024knowla} integrates KGE to activate relevant knowledge in an input question by infusing the relevant entity embeddings into the respective textual token embeddings of the input text while fine-tuning the decoder model.
Therefore, they use entity linking methods to find relevant entities for the tokens in the input question using existing methods. The mapping and infusion of KGE into LLMs is accomplished using a trainable adapter layer.

A slightly different approach is autonomous retrofitting~\cite{guan2024mitigating}, a framework that extracts, verifies, and refines factual knowledge throughout the entire reasoning process of LLMs. 
This process is entirely executed by LLMs without requiring additional external effort. 
It operates in reverse, starting from the LLM’s output, which is used to generate a query. 
Relevant facts are then retrieved from the knowledge graph based on the query, and the response is refined using these facts.

Significant effort has been devoted to incorporating KGs into the training process of LLMs~\cite{zhang2019ernie,wang2021kepler,peters2019knowledge}. 
However, a major limitation of these approaches is the inability to integrate KG updates without re-training, in addition to the high cost of the pre-training process itself.

Furthermore, substantial research has been conducted to incorporate KG in the fine-tuning stage by designing a special KG encoder of KG information.
In some approaches, pre-trained LLMs augmented with KG information are fine-tuned with full parameter updates~\cite{lin2019kagnet,yasunaga2021qa,sun2021jointlk,park2023relation}, in others, intermediate results of reasoning on KGs are used to generate a fine-tuning corpus~\cite{DBLP:conf/rulemlrr/BaldazziBCCGS23}.

While these approaches demonstrate significant achievements, they tend to be costly and poorly scalable. Moreover, they primarily focus on relevant fact retrieval, sub-graph, and small-scale graph generation techniques. Consequently, they fail to capture the complete structure and intricacies of the full knowledge graph.
Additionally, these methods often require multiple fine-tuning stages and rely heavily on integrating various external tools, increasing complexity and computational overhead, and thereby limiting scalability and practicality.
Our method addresses these limitations by aligning directly the reference entity embedding with the language model embedding space while capturing the global KG structure, meanwhile using only one simple projection layer training stage and one PEFT-based end-to-end fine-tuning stage.

\section{ALIGNed-LLM Approach}\label{sec:kge-llm}

The main goal of our work is to employ pre-trained  knowledge graph embedding models to enhance the performance of language models by biasing their output more towards the fact in the KG, thereby reducing hallucinations. 
Specifically, our primary goal is to improve the LLM ability to achieve factual correctness and to align more closely with the facts of our domain of interest, by incorporating external information from a large-scale structured source like KGs. 

\begin{figure*}[ht]
    \centering
    {\includegraphics[width=.75\textwidth]{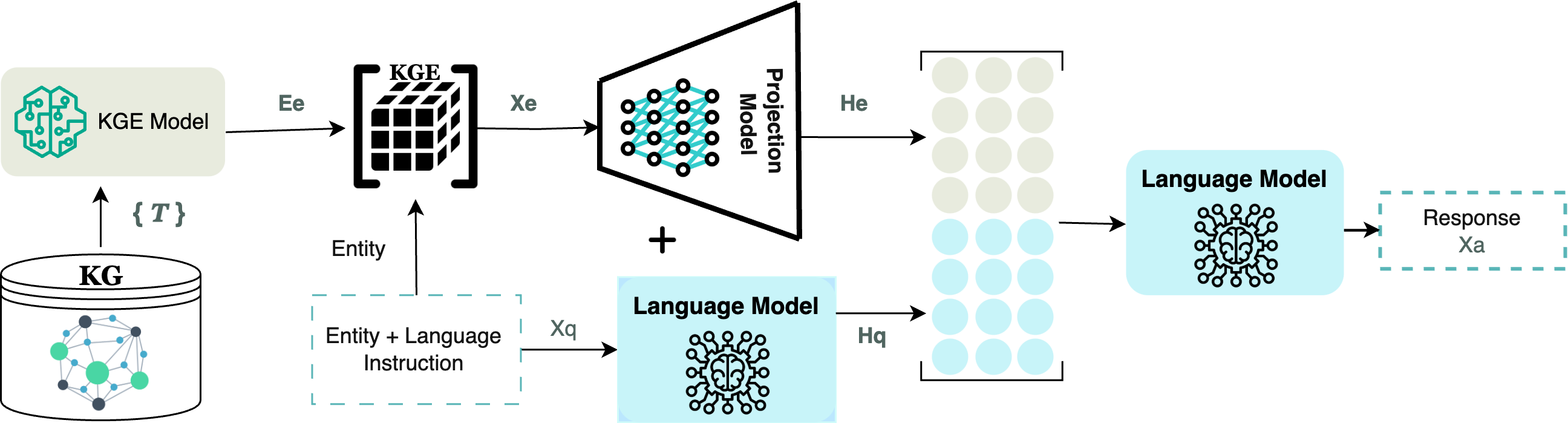}}
    \caption{ALIGNed-LLM Pipeline. Pre-trained KG embeddings (left) are used to project text embeddings for subsequent QA (right).}
    \label{fig:arch}
\end{figure*}

In order to do so, we propose ALIGNed-LLM, Adaptive Learning of Instruction-tuned Graphs and Language Models. 
The pipeline of ALIGNed-LLM is shown in Figure \ref{fig:arch}.  
We make use of a pre-trained KGE model, which contains entity embeddings. This model is frozen, i.e.~requires no updates, in our approach. 

The QA dataset consists of pairs of a reference entity and a query which is processed as per the instruction template of the LLM. In our approach, we assume that the query is constructed around a single entity $e$ that is provided along with the instruction.
The input to ALIGNed-LLM consists of a textual query X\textsubscript{q} and a reference entity. In the instruction processing step,
 
we retrieve the corresponding entity embedding X\textsubscript{e}.
Next the entity  embedding X\textsubscript{e} is projected into text embedding space using a lightweight trainable projection layer. 
After that, the transformed entity embedding H\textsubscript{e} is concatenated with the language token embedding H\textsubscript{q} of the query. 
The combined input containing both textual context and entity specific information is then passed to a decoder language model to generate the final response X\textsubscript{a}. 

\begin{figure*}[ht]
    \centering
    \includegraphics[width= 1 \textwidth]{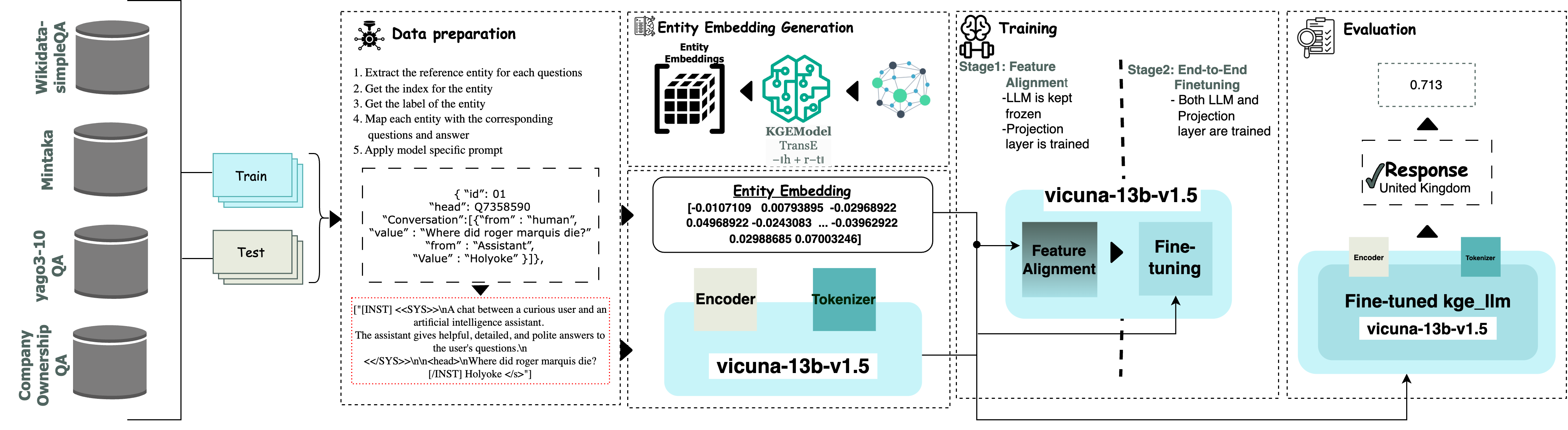}
    \caption{Overview of our approach, showing the workflow of ALIGNed-LLM model.}
    \label{fig:cs}
\end{figure*}

\subsection{Generation of Entity Embedding}
\textbf{A Knowledge Graph} is a multi-relational directed graph $ \mathcal{KG} =(\mathcal{E}, \mathcal{R}, \mathcal{T})$ where $\mathcal{E}, \mathcal{R}$ are the set of nodes (entities) and edges (relations between entities), respectively. 
The set $\mathcal{T} = \{(X_h, r, X_t) \subseteq \mathcal{E} \times \mathcal{R} \times \mathcal{E}\}$ contains all the triples of the $\mathcal{KG}$, also called positive or true facts; $h$ and $t$ are the head and the tail entities, respectively, and $r$ is the relation. 

\smallskip\noindent \textbf{A Knowledge Graph Embedding} model learns vector representations of entities ($\mathcal{E}$) and relations ($\mathcal{R}$). 
A vector representation denoted by ($\mathbf{X}_h, \mathbf{r}, \mathbf{X}_t$) is learnt by the model per triple $(X_h,r,X_t)$, where $\mathbf{X}_h,\mathbf{X}_t$ $\in$ $\mathbf{V}^{d_e}$, $\mathbf{r}$ $\in$ $\mathbf{V}^{d_r}$. $\mathbf{V}^{d}$ is a $d$-dimensional vector space, and $d_r$ and $d_e$ denote relations $r$ and entities $e$ embedding dimensions, respectively. 
KGE models can be categorized into tensor decomposition, deep learning, and geometric models according to how the interaction between relations and entities is modeled. 
Geometric models represent relations as geometric transformations from the head to the tail entities. 
We can further categorize geometric models based on whether the geometric transformation involves translation, rotation, or a combination of both \cite{cao2211knowledge}. 
TransE \cite{bordes2013translating} is the first translation-based and a state-of-the-art KGE model. 
TransE embeds entities and relations in the same real vector space. 
Relationships are exclusively modeled as translations between these vectors. Thus, TransE learns the embeddings such that
\begin{equation}
    \mathbf{X_{h}}+\mathbf{r} \approx \mathbf{X_{t}}
\end{equation}   
TransE stands out for its analytical simplicity, making it easy to implement while ensuring computational efficiency and scalability to large datasets.  
 
Motivated by these advantages, we decided to use TransE for our study. 
It is worth noting that our framework is general, and TransE can be replaced by any other KGE model. 
In fact, different models may perform better depending on the graph structure, as they can more effectively capture and learn the relational and structural patterns within the graph.

For a given QA dataset, we consider a reference knowledge graph. We trained TransE on the reference KG to learn triple embeddings and stored entity embeddings in a matrix $\mathbf{E}$ called lookup table. The entity embeddings are therefore infused with the KG structural information.
 
\subsection{Projection Layer} \label{sec:proj}
The projection layer $\phi$ is introduced to bring entity embeddings $\mathbf{X}_{e} \in \mathbb{R}^{d_{e}}$ into the text embedding space $\mathbb{R}^{d_{q}}$. 
Thus, after retrieving the entity embedding $\mathbf{X}_e$ from the lookup table, $\mathbf{X}_e$ is passed through a trainable neural network to obtain 
\begin{equation}
    \mathbf{H}_{e} = \phi(\mathbf{X}_{e}) \in \mathbb{R}^{d_{q}}.
\end{equation}
Depending on the required configuration, the projection layer is implemented as the following: 
\paragraph{Identity Map.} The projection layer is set to the identity map when the KGE embeds the entities directly into the text embedding vector space. This means
 \begin{equation}
\mathbf{H}_{e} = \mathbf{X}_{e}.
 \end{equation}
Otherwise, we consider one of the following transformations:
\paragraph{Linear Projection.} A simple linear transformation is applied on the input providing straightforward linear mapping of the entity embedding into the text embedding space. The mathematical representation of the layer is given by 
\begin{equation}
     \mathbf{H}_{e} = \mathbf{W}_{e}\mathbf{X}_{e} + \mathbf{b}_{e}
\end{equation}
where, $\mathbf{W}$ and $\mathbf{b}$ are weights and biases. 
\paragraph{Complex Projection.} It incorporates a multi-layered perceptron (MLP) with Gaussian Error Linear Unit (GELU) activation function based on the depth denoted by $n$, which determines the number of linear transformations and GELU activations. This enables more sophisticated transformations for complex alignment between entity and language model's text embedding spaces. 
We can express this transformation as 
    
\begin{align}
    \mathbf{H}_{e} &= f\left(f_n(\cdots f_2(f_1(\mathbf{X}_e))\cdots )\right)
\end{align}
where $f_k(\mathbf{X}) = \mathit{GELU}(\mathbf{W}_{k}\mathbf{X}+\mathbf{b}_{k}), k=1, \ldots, n$ and $f$ is an activation function.

\subsection{Infusion of KGE}
Once the entity and text embeddings have been aligned, the final step is to infuse both embeddings and provide them as input to the language model. 
The primary objective of this step is to provide the model with additional entity-specific context that will help it distinguish between similar types of entities---for example, between ``Berlin'' as a place and ``Berlin'' as a name. 
Thus, such entities will be further apart in the embedding space, reducing ambiguity and mitigating hallucinations. 
The integration of the representations is achieved by a tensor concatenation operation, formulated as follows:
\begin{equation}
\mathbf{H} = [\mathbf{H}_{q}: \mathbf{H}_{e}]
\end{equation}

With a probabilistic formulation, we maximize the likelihood of generating the correct response X\textsubscript{a}, given the query X\textsubscript{q} and the entity embedding X\textsubscript{e} from the KG, as follows: 
\begin{equation}
    \begin{aligned}
        P(\mathbf{X_{a}} \mid \mathbf{X{q}},\mathbf{X{e}}) &= P(\mathbf{X_{a}}| \mathbf{H}) \\
        &  = \prod_{l=1}^{L} P(\mathbf{X_a^l} \mid \mathbf{X_a^{<l}}, \mathbf{H})
    \end{aligned}
\end{equation}
where L is the length of the response sequence and $\mathbf{X_a^{<l}}$ represents the token generated up to l\textsuperscript{th} position. 
        
\subsection{Training Procedure}
The training procedure of ALINGNed-LLM follows a two-stage process consisting of (i) a \textit{feature alignment}, and (ii) an \textit{end-to-end fine-tuning.} 
In the first stage, the entity and text features are aligned by training the projection layer in such a way that the weights and biases of the language model are frozen, while only the projection matrix is trained. In this stage, the entity representations are effectively mapped into the model's latent space. 

In the second stage, the projection matrix along with the head layer of the language model are jointly trained. This process enhances the model's ability to generate accurate responses grounded in both text and knowledge graph. 
The goal of the training process is to minimize the distance between the generated response by the language model $\mathcal{L}_{\text{LM}}$ and ground truth, using a loss function, such as cross-entropy, as follows: 

\begin{equation} \label{eq:l-finetune}
    \begin{aligned}
    \mathbf{X_a} &= \mathcal{L}_{\text{LM}}(\mathbf{H}) \\
    \mathcal{L}_{\text{fine-tune}} &= \mathit{CrossEntropy}(\mathbf{X_a}, \mathit{GroundTruth})
    \end{aligned}
\end{equation}

\smallskip
\noindent The prompts used through the stages are: 
\smallskip

\noindent\textbf{Feature Alignment Stage:} \\    
\indent \texttt{System Message <STOP>}\\
\indent \texttt{Human: Entity <STOP>}\\
\indent \texttt{Assistant: Response <STOP>}
    
\smallskip\noindent\textbf{End-to-End Fine-tuning:} \\    
\indent \texttt{System Message <STOP>}\\
\indent \texttt{Human:<Entity> Instruction <STOP>}\\
\indent \texttt{Assistant: Response <STOP>}

\begin{table*}[t]\centering
    \caption{ALIGNed-LLM performance on four Question Answering datasets. The comparison metrics are: Exact Match denoted by EM, Rouge1, RougeL Reverse Weighted Bleu denoted as RWB, and F1 score. }\label{tab: r-multiple-datasets}
    \scriptsize
    \begin{tabular}{l|rrrrrr|rrrrrr}
        \toprule
             &\multicolumn{5}{c}{\cellcolor{gray!20}Wikidata} & &\multicolumn{5}{c}{\cellcolor{gray!20}YAGO} \\\midrule
        \textbf{Model} &\textbf{EM} &\textbf{Rouge1} &\textbf{RougeL} &\textbf{RWB} &\textbf{F1} & &\textbf{EM} &\textbf{Rouge1} &\textbf{RougeL} &\textbf{RWB} &\textbf{F1}
        \\\midrule
        Vicuna-13B &0.657 &0.698 &0.698 &0.708 &0.661 &&0.296 &0.393 &0.393 &0.258 &0.274 \\
        ALIGNed\_Vicuna-13B &0.664 &0.704 &0.706 &0.725 &0.669 &&0.313 &0.421 &0.421 &0.287 &0.291 \\
        \midrule
        $\Delta$ &\textcolor{teal}{\textbf{0.70\%}} &\textcolor{teal}{\textbf{0.60\%}} &\textcolor{teal}{\textbf{0.80\%}} &\textcolor{teal}{\textbf{1.70\%}} &\textcolor{teal}{\textbf{0.80\%}} &&\textcolor{teal}{\textbf{1.70\%}} &\textcolor{teal}{\textbf{2.80\%}} &\textcolor{teal}{\textbf{2.80\%}} &\textcolor{teal}{\textbf{2.90\%}} &\textcolor{teal}{\textbf{1.70\%}} 
             
        \\\midrule
        Mistral-7B &0.221 &0.287 &0.288 &0.327 &0.130 &&0.000 &0.031 &0.031 &0.008 &0.000 \\
        ALIGNed\_Mistral-7B &0.605 &0.647 &0.646 &0.679 &0.596 &&0.134 &0.178 &0.178 &0.100 &0.052 
        \\\midrule
        $\Delta$ &\textcolor{teal}{\textbf{38.40\%}} &\textcolor{teal}{\textbf{36.00\%}} &\textcolor{teal}{\textbf{35.80\%}} &\textcolor{teal}{\textbf{35.20\%}} &\textcolor{teal}{\textbf{46.60\%}} && \textcolor{teal}{\textbf{13.40\%}} &\textcolor{teal}{\textbf{14.70\%}} &\textcolor{teal}{\textbf{14.70\%}} &\textcolor{teal}{\textbf{9.20\%}} &\textcolor{teal}{\textbf{5.20\%}} 
        \\\midrule
        TL-1.1B &0.539 &0.588 &0.588 &0.620 &0.530 &&0.253 &0.353 &0.352 &0.228 &0.227 \\
        ALIGNed\_TL-1.1B &0.652 &0.693 &0.693 &0.779 &0.658 &&0.301 &0.408 &0.408 &0.273 &0.269 
        \\\midrule
        $\Delta$ &\textcolor{teal}{\textbf{11.30\%}} &\textcolor{teal}{\textbf{10.50\%}} &\textcolor{teal}{\textbf{10.50\%}} &\textcolor{teal}{\textbf{15.90\%}} &\textcolor{teal}{\textbf{12.80\%}} && \textcolor{teal}{\textbf{4.80\%}} &\textcolor{teal}{\textbf{5.50\%}} &\textcolor{teal}{\textbf{5.60\%}} &\textcolor{teal}{\textbf{4.50\%}} &\textcolor{teal}{\textbf{4.20\%}}\\
        \toprule
             &\multicolumn{5}{c}{\cellcolor{gray!20}Company Ownership} & &\multicolumn{5}{c}{\cellcolor{gray!20}Mintaka} \\\midrule
        \textbf{Model} &\textbf{EM} &\textbf{Rouge1} &\textbf{RougeL} &\textbf{RWB} &\textbf{F1} & &\textbf{EM} &\textbf{Rouge1} &\textbf{RougeL} &\textbf{RWB} &\textbf{F1}
        \\\midrule
        Vicuna-13B &0.648 &0.666 &0.666 &0.381 &0.623 &&0.576 &0.627 &0.627 &0.586 &0.567 \\
        ALIGNed\_Vicuna-13B &0.713 &0.730 &0.730 &0.484 &0.690 &&0.630 & 0.677 &0.678 &0.638 &0.623 \\
        \midrule
        $\Delta$ &\textcolor{teal}{\textbf{6.50\%}} &\textcolor{teal}{\textbf{6.40\%}} &\textcolor{teal}{\textbf{6.40\%}} &\textcolor{teal}{\textbf{10.30\%}} &\textcolor{teal}{\textbf{6.70\%}} && \textcolor{teal}{\textbf{5.40\%}} &\textcolor{teal}{\textbf{5.00\%}} &\textcolor{teal}{\textbf{5.10\%}} &\textcolor{teal}{\textbf{5.20\%}} &\textcolor{teal}{\textbf{5.60\%}}\\
        \midrule
        Mistral-7B &0.380 &0.385 &0.385 &0.145 &0.378 &&0.552 &0.605 &0.604 &0.553 &0.542 \\
        ALIGNed\_Mistral-7B& 0.488 &0.511 &0.512 &0.235 &0.473 && 0.568 &0.618 &0.617 &0.587 &0.557 \\
        \midrule
        $\Delta$ &\textcolor{teal}{\textbf{10.80\%}} &\textcolor{teal}{\textbf{12.60\%}} &\textcolor{teal}{\textbf{12.70\%}} &\textcolor{teal}{\textbf{9.00\%}} &\textcolor{teal}{\textbf{9.50\%}} && \textcolor{teal}{\textbf{1.60\%}} &\textcolor{teal}{\textbf{1.30\%}} &\textcolor{teal}{\textbf{1.30\%}} &\textcolor{teal}{\textbf{3.40\%}} &\textcolor{teal}{\textbf{1.50\%}}\\
        \midrule
        TL-1.1B &0.766 &0.779 &0.779 &0.471 &0.733 &&0.389 &0.449 &0.447 &0.425 &0.366 \\
        ALIGNed\_TL-1.1B& 0.901 &0.916 &0.915 &0.789 &0.897 && 0.402 &0.462 &0.462 &0.43 &0.380 \\
        \midrule
        $\Delta$ &\textcolor{teal}{\textbf{13.50\%}} &\textcolor{teal}{\textbf{13.70\%}} &\textcolor{teal}{\textbf{13.60\%}} &\textcolor{teal}{\textbf{31.80\%}} &\textcolor{teal}{\textbf{16.40\%}} && \textcolor{teal}{\textbf{1.30\%}} &\textcolor{teal}{\textbf{1.30\%}} &\textcolor{teal}{\textbf{1.50\%}} &\textcolor{teal}{\textbf{0.50\%}} &\textcolor{teal}{\textbf{1.40\%}}\\
        \bottomrule
    \end{tabular}
        \label{tab:performance}
\end{table*}

\section{Experiments}
\label{sec:experiments}
This section evaluates the performance and effectiveness of our framework in addressing factuality of LLMs.
Additional results and implementation details are provided in the supplementary material. 

\textbf{Question Answering Task.} The QA task (see Figure~\ref{fig:cs}) involves answering natural language questions using factually correct and structured data. 
Each question is associated with a specific entity (\textit{head entity}) and its relation to the answer (\textit{tail entity}). 
Formally, the task is to predict the tail entity $X_t$ given a question $X_q$ and a head entity $X_h$ using the relation $r$. QA datasets are structured as tuples $(X_h, Xq, X_t)$, where $Xq$ is formulated around $X_h$ and $r$.
This enables reasoning over the KG to find $X_t$, or vice versa.

\smallskip\noindent\textbf{Evaluation Metrics}. We evaluate the performance of our framework by measuring the percentage of correct answers on the QA test datasets. 
The metrics that are adopted for the analysis are: ROUGE-L, which evaluates the alignment of generated text with the reference by considering both content coverage (recall) and efficiency of generation (precision); and ROUGE-1 that refers to the overlap of unigrams.
We also use the BLEU metric, which evaluates text generation by measuring n-gram precision compared to a reference. 
Specifically, we consider the Reverse Weighted BLEU (RWB) where lower-order n-grams (in our case BLEU-1 to BLEU-4) are prioritized over higher-order ones to emphasize content coverage.
We also report on F1 score that balances between recall and precision as well as Exact Match (EM) to measure the percentage of outputs that exactly match the reference text.

\smallskip\noindent\textbf{Baselines Models and Datasets}.
We consider the following open-source LLMs: \textit{TinyLlama-1.1B (TL-1.1B)}~\cite{zhang2024tinyllama}, \textit{vicuna-13b-v1.5 (Vicuna-13B)}~\cite{zheng2023judging}, 
    \textit{Mistral-7B-Instruct-v0.3(Mistral-7B)}~\cite{jiang2023mistral}. 
To evaluate the exact improvement introduced by our ALIGNed-LLM framework, we consider as baselines the fine-tuned version of each LLM. 
The fine-tuning uses the textual QA pairs extracted from the KG to ensure a fair comparison, i.e., it guarantees that the baseline models possess the same knowledge as our ALIGNed models. 
In the tables, we refer to our solution using the prefix \textit{ALIGNed-'LLM Name'}.

\begin{table}[t]
\centering
\caption{Dataset Statistics. This table presents the number of distinct entities, relations, division of training and test sets, and pre-trained TransE MRR score on corresponding KGs. The dimension used for TransE for YAGO-310 and CO KGs is 1024, while for Wikidata5M it is 512. AWC stands for average word count for answers.}
\label{tab:data}
\begin{adjustbox}{width=0.9\columnwidth}
\begin{tabular}{lrrrrrrr}\toprule
\textbf{Dataset} & $|E|$ & $|R|$ & \textbf{Train Size} & \textbf{Test Size} & \textbf{AWC} & \textbf{TransE\_MRR} \\\midrule
YAGO3-10 &112981 &37 &101683 &11298 &2.185 &0.515 \\
Mintaka &4893 &- &11298 &3702 &1.875 &0.253 \\
Wikidata &18006 &129 &22430 &633 &1.9 &0.253 \\
CO &14720 &7 &117074 &13008 &1.68 &0.570 \\

\bottomrule
\end{tabular}
\end{adjustbox}
\end{table}

\smallskip

We used four QA datasets constructed using 3 KGs.The quantitative details are shown in Table~\ref{tab:data}, where the result for Wikidata5m is from~\cite{wang2021kepler}.

Wikidata5M \cite{wang2021kepler} is a large-scale KG that contains 5 million entities spanning the general domain. 
For Wikidata5M  KG, we considered the following two QA datasets: \textbf{(1) Wikidata QA dataset}~\cite{banerjee2023gett} - is constructed using the KG triples directly, where the questions refer to the head entity and a specific relation, and the tail entity serves as the answer. 
\textbf{(2) Mintaka QA dataset}~\cite{sen2022mintaka} - is a complex and multilingual dataset with different levels of complexity such as count, comparative, superlative, ordinal, multi-hop, etc. 
YAGO3-10 is a subset of the broader YAGO3  KG~\cite{yago3_10_dataset}, which integrates generic information from Wikipedia and WordNet.  
It includes only entities that are connected by at least 10 distinct relations. 
We generated the \textbf{(3) YAGO310 QA} dataset using the question template based on 37 relations. 

The \textbf{(4) Company Ownership (CO)} QA based on CO KG. This anonymised KG from the central bank captures the relationships between companies: nodes represent individuals or companies, and edges are the shareholding links between them, such as ownership, control, ultimate control (as defined in Section~\ref{sec:intro}), and so on.
The graph comprises approximately 1,657 M weakly connected components and its degree distribution follows a scale-free power-law, as common in corporate economics settings.
Our dataset is an anoynmized representative sub-graph of the CO KG. 
It consists of around 29,318 distinct entities, from which we extracted around 100k triples. 
We built the CO dataset considering two structured QA templates of increasing complexity, for each of the 7 types of relationships in the KG. \\

As for obtaining the embeddings of the knowledge in these datasets, we used the pre-trained TransE model for Wikidata presented in GraphVite~\cite{zhu2019graphvite}.
For all the other datasets (YAGO, Mintaka, and CO), we trained the TransE model presented in the RotatE framework~\cite{sun2019rotate}.

\subsection{Results and Analysis}
We present experimental results showing that aligning LLMs with the factual accuracy of KGs significantly improves their precision.
Table \ref{tab:performance} provides an extensive comparison of ALIGNed-LLM across three benchmark datasets, as well as the proprietary --- anonymised --- dataset of company ownerships (CO).

According to the EM metric, the results show a consistent and significant improvement across all datasets when using the ALIGNed approach. 
For instance, significant gains are observed with ALIGNed\_Mistral-7B (38.40\%) and ALIGNed\_TL-1.1B (11.30\%), while Wikidata also sees a slight improvement of 0.70\% with ALIGNed\_Vicuna-13B. 
Similarly, for the Company Ownership dataset, ALIGNed\_TL-1.1B achieves a significant $\Delta$ of 13.50\% in comparison to TL-1.1B alone, highlighting the model's capacity to improve precision in identifying exact answers.
Also in the case of both the Rouge1 and RougeL metrics, the ALIGNed language models show consistent improvement across all datasets. Notably, Rouge1 for the YAGO dataset improves by 14.70\% for ALIGNed\_Mistral-7B and 5.50\% for ALIGNed\_TL-1.1B, showcasing the robustness of the approach in capturing key content. 
With RougeL, the Mistral-7B model using our method (ALIGNed\_Mistral-7B) gains around 35.80\%, and the ALIGNed\_TL-1.1B model shows a significant improvement of 10.50\% for Wikidata.
These results with ALIGNed-LLM emphasize an enhanced structural alignment of the generated and reference answers.
\begin{figure}[t]
    \centerline{\includegraphics[width=.9\columnwidth]{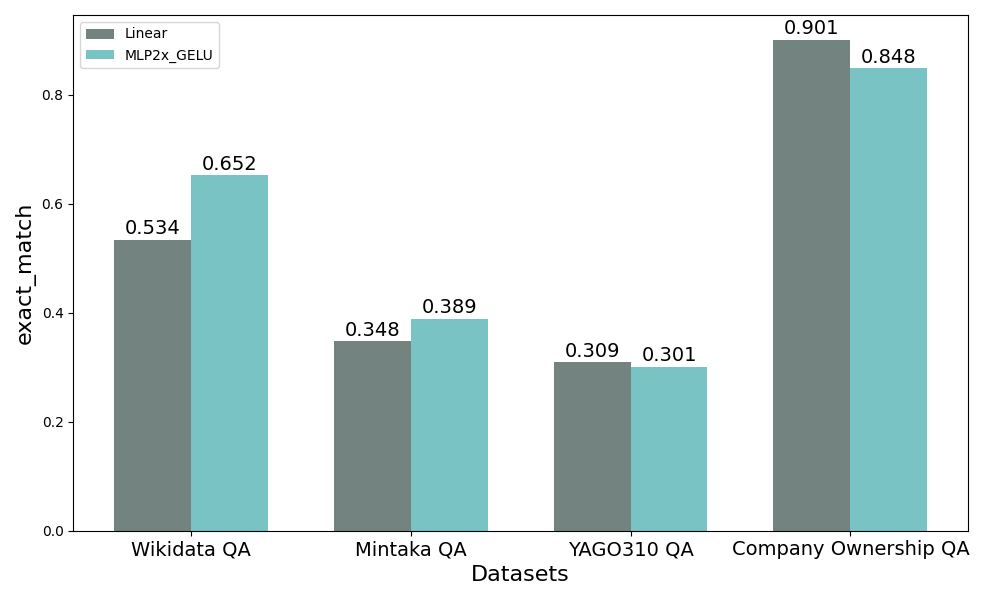}}
    \caption{Impact of Linear vs. Complex Projections across Datasets.}
    \label{fig:proj-bar}
\end{figure}

\begin{figure}[t]
\centerline{\includegraphics[width=.9\columnwidth]{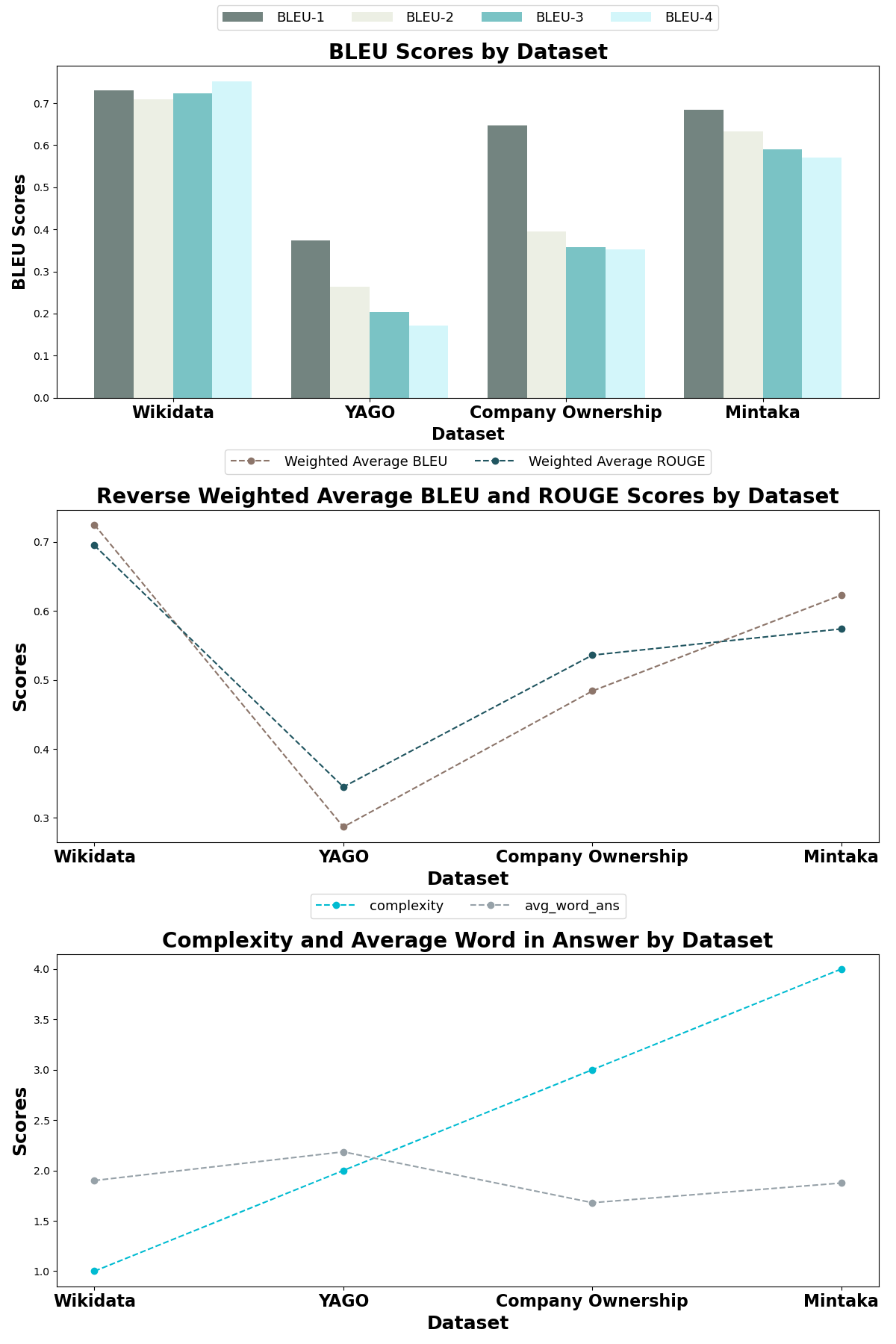}}
    \caption{ Evaluation metric analysis of ALIGNed\_Vicuna-13B model on all datasets.}
    \label{fig:metric-analysis}
\end{figure}
With the Wikidata dataset, the RWB metrics show the largest relative improvement for ALIGNed\_Vicuna-13B (1.70\%) and ALIGNed\_Mistral-7B (35.20\%). 
The 46.60\% gain in the F1 score in this dataset for ALIGNed\_Mistral-7B further reinforces the better alignment of precision and recall of the underlying language model after using the proposed approach to generate more accurate responses.
For the YAGO dataset, we observed a moderate improvement across all metrics for ALIGNed\_Vicuna-13B, for example  2.90\% in RWB, and 2.80\% in RougeL. 
However, more substantial improvements, such as 14.70\% in Rouge1 and  13.40\% in EM, are shown by the ALIGNed\_Mistral-7B model.
In the CO dataset, significant improvements are seen in both EM and RWB, with ALIGNed\_TL-1.1B achieving 13.50\% in EM and 31.80\% in RWB.
The results indicate the adaptability of the ALIGNed to datasets with different entity-relation structures.

For the Mintaka dataset, we notice moderate improvement for all models. RWB metrics depict modest improvement of 3.40\% with ALIGNed\_Mistral-7B while relative improvement for  F1 score for ALIGNed\_Vicuna-13B is 5.60\%. 

Therefore, these results provide a clear and transparent evaluation, emphasizing that the ALIGNed approach demonstrates its advantages and limitations across various datasets. We also employed T-Test considering metrics of Rouge[1,2,L,Lsum], and Bleu[1-4].
Out of a total of 24 tests, 22 were found to be statistically significant. 

\smallskip\noindent\textbf{Effect of Projection Type}. 
In Figure \ref{fig:proj-bar}, we show the difference of using linear vs.\ complex projection on ALIGNed\_TL-1.1B model over all datasets. 
We chose EM metric to showcase the impact here. 
The model gains better results on other datasets than CO, when using the complex projection layer with the GELU activation function. 
The 5.3\% improvement in the CO dataset with the linear projection vs.\ the complex one can be attributed to the structure of the CO KG (more sparse than other KGs). 

Thus we observe that the choice of the best projection strategy depends on the KG topology and structure.

\smallskip\noindent\textbf{Correlation of data and Measurements}. 
Although BLEU measures the challenge of aligning increasingly long phrasing, its effectiveness varies on the basis of the complexity of the alignment being measured.

BLEU-1, which emphasizes unigram precision, often scores higher as it captures individual word matches; BLEU-4, requiring longer n-gram alignment, is more sensitive to structural coherence, sentence complexity, and sequence length. Similar considerations can be developed for the ROUGE scores. In Figure~\ref{fig:metric-analysis}, we analyzed this relationship in terms of the complexity and average word count of the answers across datasets.
The high complexity of CO causes a visible drop in BLEU-2, and only a moderate drop in BLEU-3 and BLEU-4, with a good average BLEU and ROUGE scores. Less complex cases, like Wikidata are less affected by this phenomenon. Finally, despite the high complexity, with Mintaka, high BLEU scores are observed, with a drop in BLEU-4 that is mitigated by the low word count and complexity of the dataset, as discussed earlier.
\section{Conclusion}
\label{sec:conclusion}
We introduced ALIGNed-LLM, a framework integrating Knowledge Graph Embeddings (KGE) into LLMs, aligning entity and text embeddings via a trainable projection layer.
ALIGNed-LLM enhances factual accuracy and reasoning of LLMs which mitigate their hallucinations, entity ambiguity, and knowledge gaps. Our approach bridges probabilistic inference with grounded knowledge. The effectiveness of ALIGNed-LLM is demonstrated by extensive experiments across diverse datasets. Its potential for practical use is highlighted by its notable adaptability to domain-specific knowledge and ease of knowledge update (requiring only embedding updates). ALIGNed-LLM will be broadened to handle temporal and multi-modal knowledge graphs to enable processing and integration information across different time points and various modalities such as text, images, and structured data thus expanding its applicability.

\section*{Ethical Statement} The Company Ownership dataset was processed in compliance with ethical standards and relevant laws, including data protection regulations (e.g., GDPR). All data were anonymized and analyzed within secure systems. No personally identifiable information was accessed or disclosed.
The other datasets are publicly available and do not contain any personal data. 

\printbibliography

\clearpage
\appendix

\section{Experimental details}

\subsection{Question-Answering dataset}
The anonymized company ownership question-answering dataset is built according the following two templates in Table~\ref{tab:company_own_qa} and Table~\ref{tab:company_own_qa_reasoning}, which consists of increasingly complexity queries. For each query we build one positive example, i.e., consists of $<$query, correct answer$>$, and a negative example, i.e., $<$query, wrong answer$>$. We removed entities where the answers containing more than 20 entities from the dataset, to avoid bottlenecks and limitations for the used LLMs. 

\begin{table}[ht!]
    \centering
    \small
\resizebox{0.4\textwidth}{!}{
    \begin{tabular}{|p{2cm}|p{8cm}|}
        \hline
        \textbf{Category} & \textbf{Query Templates} \\ \hline
        \textbf{Ultimate Control} & 
        \begin{itemize}
            \item Who is the ultimate controller of \{Y\}?
            \item Who holds ultimate control over \{Y\}?
            \item Which company or person is the ultimate controller of \{Y\}?
            \item Is it true that \{X\} is the ultimate controller of \{Y\}?
            \item Does \{X\} have ultimate control over \{Y\}?
            \item Can we confirm that \{X\} is the top controller of \{Y\}?
        \end{itemize} \\ \hline
        \textbf{Control} & 
        \begin{itemize}
            \item Who controls \{Y\}?
            \item Who is the controller of \{Y\}?
            \item Which company or person has control over \{Y\}?
            \item Does \{X\} control \{Y\}?
            \item Is it true that \{X\} has control over \{Y\}?
            \item Can we verify that \{X\} is a controller of \{Y\}?
        \end{itemize} \\ \hline
        \textbf{Ownership} & 
        \begin{itemize}
            \item Who owns \{Y\}?
            \item Who is the owner of \{Y\}?
            \item Which entity is the owner of \{Y\}?
            \item Is it true that \{X\} owns shares of \{Y\}?
            \item Does \{X\} have ownership over \{Y\}?
            \item Can we confirm that \{X\} owns shares in \{Y\}?
        \end{itemize} \\ \hline
        \textbf{Role} & 
        \begin{itemize}
            \item Who has a role in \{Y\}?
            \item Who takes a role in \{Y\}?
            \item Which company or person assumes a role with \{Y\}?
            \item Does \{X\} have a role in \{Y\}?
            \item Is it true that \{X\} is assigned a role in \{Y\}?
            \item Can we confirm that \{X\} plays a role in \{Y\}?
        \end{itemize} \\ \hline
        \textbf{Qualified Holdings} & 
        \begin{itemize}
            \item Who has qualified holdings in \{Y\}?
            \item Who is a qualified holder of \{Y\}?
            \item Which company or person holds a qualified position in \{Y\}?
            \item Is it true that \{X\} has qualified holdings in \{Y\}?
            \item Does \{X\} qualify as a holder in \{Y\}?
            \item Can we confirm that \{X\} is a qualified holder of \{Y\}?
        \end{itemize} \\ \hline
        \textbf{Reachability} & 
        \begin{itemize}
            \item Is it true that \{X\} is connected to \{Y\}?
            \item Can \{X\} reach \{Y\} through any connections?
            \item Does \{X\} have a connection to \{Y\}?
            \item Is it possible for \{X\} to be accessed by \{Y\}?
        \end{itemize} \\ \hline
        \textbf{Influence} & 
        \begin{itemize}
            \item Who influences \{Y\}?
            \item Who has influence over \{Y\}?
            \item Is it true that \{X\} influences \{Y\}?
            \item Does \{X\} have influence on \{Y\}?
            \item Can \{X\} influence \{Y\}?
            \item Is there evidence that \{X\} exerts influence over \{Y\}?
        \end{itemize} \\ \hline
    \end{tabular}}
    \caption{Company Ownership Q/A.}\label{tab:company_own_qa}
\end{table}

\begin{table}[h!]
    \centering
    \small
\resizebox{0.4\textwidth}{!}{
    \begin{tabular}{|p{2cm}|p{8cm}|}
        \hline
        \textbf{Category} & \textbf{Query Templates} \\ \hline
        \textbf{Control} & 
        \begin{itemize}
            \item How many companies and/or people control \{Y\}?
            \item Is it true that \{Y\} is controlled by \{Z\} companies and/or people?
        \end{itemize} \\ \hline
        \textbf{Ownership} & 
        \begin{itemize}
            \item How many companies own \{Y\}?
            \item Is it true that \{Y\} is owned by \{Z\} companies and/or people?
        \end{itemize} \\ \hline
        \textbf{Role} & 
        \begin{itemize}
            \item How many entities have a role in \{Y\}?
            \item Is it true that \{Y\} has \{Z\} entities assuming a role in it?
        \end{itemize} \\ \hline
        \textbf{Qualified Holdings} & 
        \begin{itemize}
            \item How many entities have qualified holdings in \{Y\}?
            \item Is it true that \{Y\} has \{Z\} entities with qualified holdings?
        \end{itemize} \\ \hline
        \textbf{Reachability} & 
        \begin{itemize}
            \item How many entities can reach \{Y\} through any connections?
            \item Is it true that \{Y\} is connected with \{Z\} companies and/or people?
        \end{itemize} \\ \hline
        \textbf{Influence} & 
        \begin{itemize}
            \item How many companies and/or people influence \{Y\}?
            \item Is it true that \{Y\} is influenced by \{Z\} companies and/or people?
        \end{itemize} \\ \hline
    \end{tabular}}
    \caption{Company Ownership reasoning Q/A.}\label{tab:company_own_qa_reasoning}
\end{table}

The YAGO3-10 QA dataset is constructed using the query template shown in Tabale \ref{tab-yago-qt}. The queries we formulated considering the 37 relationships in YAGO3-10 KG. For each relation group, the corresponding \{x\} value is replaced by respective entities in the triples.

\begin{table*}[h!]
    \centering
    \caption{YAGO3-10 QA Query template.}\label{tab:yago-qa-query}
    \resizebox{0.7\textwidth}{!}{
    \begin{tabular}{|l|p{8cm}|}
        \toprule \hline
        \textbf{Relation} & \textbf{Query Templates} \\ \hline
        wasBornIn & Where was \{x\} born? \\
        \hline
        worksAt & Where does \{x\} work? \\ 
                & \{x\} is employed at which organization? \\
        \hline
        happenedIn & In which location did \{x\} occur? \\
        \hline
        isLeaderOf & Who/what is led by \{x\}? \\
        \hline
        hasWonPrize & Which prize has \{x\} won? \\
        \hline
        isAffiliatedTo & With whom is \{x\} affiliated? \\
        \hline
        hasChild & Who is a child of \{x\}? \\ 
                 & Which person is \{x\} a parent to? \\
        \hline
        imports & Which items are imported by \{x\}? \\
        \hline
        exports & Which items are exported by \{x\}? \\
        \hline
        isLocatedIn & Where is \{x\} located? \\
        \hline
        hasCurrency & What is the currency of \{x\}? \\
        \hline
        created & What was created by \{x\}? \\
        \hline
        graduatedFrom & From which institution did \{x\} graduate? \\ 
                      & \{x\} is an alumnus of which university/school? \\
        \hline
        actedIn & \{x\} acted in which movie/show? \\ 
                & Which film/show featured \{x\}? \\
        \hline
        hasGender & What is the gender of \{x\}? \\ 
                  & \{x\} identifies as which gender? \\
        \hline
        hasCapital & What is the capital of \{x\}? \\
        \hline
        isInterestedIn & What is \{x\} interested in? \\
        \hline
        isConnectedTo & What is \{x\} connected to? \\
        \hline
        influences & Who/what is influenced by \{x\}? \\
        \hline
        owns & What is owned by \{x\}? \\
        \hline
        isKnownFor & What is \{x\} known for? \\ 
                   & \{x\} is famous for what? \\
        \hline
        directed & What did \{x\} direct? \\
        \hline
        diedIn & Where did \{x\} die? \\
        \hline
        dealsWith & What does \{x\} deal with? \\
        \hline
        livesIn & Where does \{x\} live? \\
        \hline
        isPoliticianOf & Which country/region is \{x\} a politician of? \\ 
                       & \{x\} is politically active in which place? \\
        \hline
        hasWebsite & \{x\} has which website? \\
        \hline
        hasNeighbor & What is the neighboring country of \{x\}? \\ 
                    & Which country shares a border with \{x\}? \\
        \hline
        hasOfficialLanguage & What is the official language of \{x\}? \\
        \hline
        isMarriedTo & Who is \{x\} married to? \\ 
                    & \{x\} is the spouse of whom? \\
        \hline
        hasAcademicAdvisor & Who is the academic advisor of \{x\}? \\
        \hline
        isCitizenOf & Which country is \{x\} a citizen of? \\
        \hline
        edited & Which work did \{x\} edit? \\
        \hline
        playsFor & Which team does \{x\} play for? \\ 
                 & \{x\} is a member of which sports team? \\
        \hline
        participatedIn & In which war/battle/event did \{x\} participate? \\
        \hline
        wroteMusicFor & Which production did \{x\} write music for? \\ 
                      & \{x\} composed music for which work? \\
        \hline
        hasMusicalRole & What is the musical role of \{x\}? \\
        \hline
        \bottomrule
    \end{tabular}}
    \label{tab-yago-qt}
    
\end{table*}

\subsection{Implementation Details.} 
For all models and datasets, we employed a learning rate of 2e-4 for overall model training and 2e-5 for projection layer training. Additionally to optimize the learning process, we adopted a warm\_up ratio of 0.03 to gradually adjust the learning rate. For Wikidata5M and Mintaka dataset we used 100 epoch to train the projection layer whereas for YAGO-310 and Company Ownership dataset we used 50 epoch to train the projection layer. 
On the other hand, for end-to-end fine-tuning, we used a single epoch across all models and datasets to avoid the risk of over-fitting and knowledge degradation. 
Moreover, for fine-tuning the combination of LoRA and PEFT technique was used. 
All experiments were conducted on NVIDIA A100-SXM4 Tensor Core-GPUs with 40 GB HBM2 memory. During training and inference, the maximum GPU usage was 64GB GPU, and the system utilized up to 128 GB of RAM. 

\subsection{Result and Analysis}
    Table \ref{tab: r-wiki}-\ref{tab: r-min} reports detailed results of the experiments conducted on ALIGNed-LLM using 3 LLMs on 4 datasets. They also consist of a comparison study with the fine-tuned baselines on the same datasets for a fair comparison. We evaluated our model using multiple metrics to verify the consistency of our results. Table \ref{tab: r-wiki} demonstrates consistent performance improvement of our model on the Wikidata QA  dataset for all three LLMs.  Similar performance is also observed on the YAGO3-10 QA dataset in Table \ref{tab:r-yago}. Except for the BLEU-4 metric, improved performance is also observed on the Company Ownership QA dataset in Table \ref{tab:r-co }. On the contrary, we observe mixed results on the Mintaka dataset in Table \ref{tab: r-min} indicating smaller models required further optimization on complex datasets. 

 \begin{table*}[!htp]\centering
        \caption{ALIGNed-LLM performance on Wikidata QA dataset}\label{tab: r-wiki}
        \scriptsize
            \begin{tabular}{lrrrrrrrrrrr}\toprule
                \textbf{Model} &\textbf{F1} &\textbf{Rouge1} &\textbf{Rouge2} &\textbf{RougeL} &\textbf{RougeLsum} &\textbf{BLEU-1} &\textbf{BLEU-2} &\textbf{BLEU-3} &\textbf{BLEU-4} &\textbf{EM} \\\midrule
                Vicuna-13B &0.567 &0.661 &0.327 &0.698 &0.698 &0.716 &0.689 &0.707 &0.734 &0.657 \\
                ALIGNed\_Vicuna-13B &0.669 &0.704 &0.337 &0.706 &0.706 &0.731 &0.709 &0.723 &0.752 &0.664 \\
                \\\midrule
                Gain &\textcolor{teal}{\textbf{0.80\%}} &\textcolor{teal}{\textbf{0.60\%}} &\textcolor{teal}{\textbf{1.0\%}} &\textcolor{teal}{\textbf{0.80\%}} &\textcolor{teal}{\textbf{0.80\%}} &\textcolor{teal}{\textbf{1.50\%}} &\textcolor{teal}{\textbf{2.00\%}} &\textcolor{teal}{\textbf{1.60\%}} &\textcolor{teal}{\textbf{1.80\%}} &\textcolor{teal}{\textbf{0.70\%}} \\
                \\\midrule
                Mistral-7B &0.130 &0.287 &0.171 &0.288 &0.289 &0.333 &0.319 &0.330 &0.320 &0.221 \\
                ALIGNed\_Mistral-7B &0.596 &0.647 &0.324 &0.646 &0.647 &0.675 &0.660 &0.692 &0.721 &0.605 \\
                \\\midrule
                Gain &\textcolor{teal}{\textbf{46.6\%}} &\textcolor{teal}{\textbf{36.0\%}} &\textcolor{teal}{\textbf{15.3\%}} &\textcolor{teal}{\textbf{35.8\%}} &\textcolor{teal}{\textbf{35.8\%}} &\textcolor{teal}{\textbf{34.2\%}} &\textcolor{teal}{\textbf{34.1\%}} &\textcolor{teal}{\textbf{36.2\%}} &\textcolor{teal}{\textbf{40.1\%}} &\textcolor{teal}{\textbf{38.4\%}} \\
                \\\midrule
                TL-1.1B &0.530 &0.588 &0.292 &0.588 &0.588 &0.620 &0.603 &0.635 &0.644 &0.539 \\
                ALIGNed\_TL-1.1B &0.658 &0.693 &0.338 &0.693 &0.692 &0.743 &0.756 &0.834 &0.882 &0.652 \\
                \\\midrule
                Gain &\textcolor{teal}{\textbf{12.8\%}} &\textcolor{teal}{\textbf{10.5\%}} &\textcolor{teal}{\textbf{4.6\%}} &\textcolor{teal}{\textbf{10.5\%}} &\textcolor{teal}{\textbf{10.4\%}} &\textcolor{teal}{\textbf{12.3\%}} &\textcolor{teal}{\textbf{15.3\%}} &\textcolor{teal}{\textbf{19.9\%}} &\textcolor{teal}{\textbf{23.8\%}} &\textcolor{teal}{\textbf{11.3\%}} \\
                \bottomrule
            \end{tabular}
        \end{table*}
        \begin{table*}[!htp]\centering
            \caption{ALIGNed-LLM performance on YAGO310 QA Dataset}\label{tab:r-yago }
            \scriptsize
                \begin{tabular}{lrrrrrrrrrrr}\toprule
                    \textbf{Model} &\textbf{F1} &\textbf{Rouge1} &\textbf{Rouge2} &\textbf{RougeL} &\textbf{RougeLsum} &\textbf{BLEU-1} &\textbf{BLEU-2} &\textbf{BLEU-3} &\textbf{BLEU-4} &\textbf{EM} \\\midrule
                    Vicuna-13B & 0.274&	0.393&	0.149&	0.393&	0.392&	0.343&	0.232&	0.182&	0.152&	0.296\\
                    ALIGNed\_Vicuna-13B & 0.291 &0.421 &0.169 &0.421 &0.421 &0.374 &0.263 &0.204 &0.172 &0.313 \\
                    \\\midrule
                     Gain &\textcolor{teal}{\textbf{1.70\%}} &\textcolor{teal}{\textbf{2.80\%}} &\textcolor{teal}{\textbf{2.0\%}} &\textcolor{teal}{\textbf{2.80\%}} &\textcolor{teal}{\textbf{2.90\%}} &\textcolor{teal}{\textbf{3.10\%}} &\textcolor{teal}{\textbf{3.10\%}} &\textcolor{teal}{\textbf{2.2\%}} &\textcolor{teal}{\textbf{2.00\%}} &\textcolor{teal}{\textbf{1.70\%}} \\
                    \\\midrule
                    Mistral-7B &  0.000 &0.031 &0.002 &0.031 &0.031 &0.019 &0.001 &0.000 &0.000 &0.000 \\
                    ALIGNed\_Mistral-7B &  0.052 &0.178 &0.065 &0.178 &0.177 &0.153 &0.082 &0.051 &0.048 &0.134 \\
                    \\\midrule
                    Gain & \textcolor{teal}{\textbf{5.2\%}} &\textcolor{teal}{\textbf{14.7\%}} &\textcolor{teal}{\textbf{6.3\%}} &\textcolor{teal}{\textbf{14.7\%}} &\textcolor{teal}{\textbf{14.6\%}} &\textcolor{teal}{\textbf{13.4\%}} &\textcolor{teal}{\textbf{8.1\%}} &\textcolor{teal}{\textbf{5.1\%}} &\textcolor{teal}{\textbf{4.8\%}} &\textcolor{teal}{\textbf{13.4\%}} \\
                    \\\midrule
                    TL-1.1B & 0.227 &0.353 &0.119 &0.352 &0.353 &0.308 &0.200 &0.156 &0.134 &0.253 \\
                    ALIGNed\_TL-1.1B & 0.269 &0.408 &0.153 &0.408 &0.407 &0.363 &0.249 &0.188 &0.158 &0.301 \\
                    \\\midrule
                    Gain & \textcolor{teal}{\textbf{4.2\%}} &\textcolor{teal}{\textbf{5.5\%}} &\textcolor{teal}{\textbf{3.4\%}} &\textcolor{teal}{\textbf{5.6\%}} &\textcolor{teal}{\textbf{5.4\%}} &\textcolor{teal}{\textbf{5.5\%}} &\textcolor{teal}{\textbf{4.9\%}} &\textcolor{teal}{\textbf{3.2\%}} &\textcolor{teal}{\textbf{2.4\%}} &\textcolor{teal}{\textbf{4.8\%}} \\
                    \bottomrule
                \end{tabular}
        \end{table*}

        \begin{table*}[!htp]\centering
            \caption{ALIGNed-LLM performance on Company Ownership QA Dataset. Notably complex projection type is used for all except ALIGNed\_TL-1.1B on CO dataset where a linear projection type is used to achieve best result. }\label{tab:r-co }
            \scriptsize
                \begin{tabular}{lrrrrrrrrrrr}\toprule
                    \textbf{Model} &\textbf{F1} &\textbf{Rouge1} &\textbf{Rouge2} &\textbf{RougeL} &\textbf{RougeLsum} &\textbf{BLEU-1} &\textbf{BLEU-2} &\textbf{BLEU-3} &\textbf{BLEU-4} &\textbf{EM} \\\midrule
                    Vicuna-13B & 0.623&	0.666&	0.042&	0.666&	0.666&	0.597&	0.233&	0.179&	0.367&	0.648 \\
                    ALIGNed\_Vicuna-13B & 0.690&	0.730&	0.084&	0.730&	0.730&	0.647&	0.395&	0.358&	0.353&	0.713 \\
                    \\\midrule
                     Gain &\textcolor{teal}{\textbf{6.7\%}} &\textcolor{teal}{\textbf{6.4\%}} &\textcolor{teal}{\textbf{4.2\%}} &\textcolor{teal}{\textbf{6.4\%}} &\textcolor{teal}{\textbf{6.4\%}} &\textcolor{teal}{\textbf{5.0\%}} &\textcolor{teal}{\textbf{16.2\%}} &\textcolor{teal}{\textbf{17.9\%}} &\textcolor{Maroon}{\textbf{-1.4\%}} &\textcolor{teal}{\textbf{6.5\%}} \\
                    \\\midrule
                    Mistral-7B & 0.378&	0.385&	0.000&	0.385&	0.385&	0.310&	0.045&	0.031&	0.012&	0.380 \\
                    ALIGNed\_Mistral-7B & 0.473&	0.511&	0.020&	0.512&	0.511&	0.436&	0.125&	0.077&	0.076&	0.488 \\
                    \\\midrule
                    Gain &\textcolor{teal}{\textbf{9.5\%}} &\textcolor{teal}{\textbf{12.6\%}} &\textcolor{teal}{\textbf{2.0\%}} &\textcolor{teal}{\textbf{12.7\%}} &\textcolor{teal}{\textbf{12.6\%}} &\textcolor{teal}{\textbf{12.6\%}} &\textcolor{teal}{\textbf{8.0\%}} &\textcolor{teal}{\textbf{4.6\%}} &\textcolor{teal}{\textbf{6.4\%}} &\textcolor{teal}{\textbf{10.8\%}} \\
                    \\\midrule
                    TL-1.1B & 0.733&	0.779&	0.062&	0.779&	0.778&	0.675&	0.332&	0.337&	0.337&	0.766 \\
                    ALIGNed\_TL-1.1B & 0.897&	0.916&	0.174&	0.915&	0.915&	0.860&	0.770&	0.726&	0.682&0.901\\
                    \\\midrule
                    Gain &\textcolor{teal}{\textbf{16.4\%}} &\textcolor{teal}{\textbf{13.7\%}} &\textcolor{teal}{\textbf{11.2\%}} &\textcolor{teal}{\textbf{13.6\%}} &\textcolor{teal}{\textbf{13.7\%}} &\textcolor{teal}{\textbf{18.5\%}} &\textcolor{teal}{\textbf{43.8\%}} &\textcolor{teal}{\textbf{38.9\%}} &\textcolor{teal}{\textbf{34.5\%}} &\textcolor{teal}{\textbf{13.5\%}} \\
                    \bottomrule
                \end{tabular}
        \end{table*}
        \begin{table*}[!htp]\centering
                \caption{ALIGNed-LLM performance on Mintaka dataset}\label{tab: r-min}
                \scriptsize
                \begin{tabular}{lrrrrrrrrrrr}\toprule
                \textbf{Model} &\textbf{F1} &\textbf{Rouge1} &\textbf{Rouge2} &\textbf{RougeL} &\textbf{RougeLsum} &\textbf{BLEU-1} &\textbf{BLEU-2} &\textbf{BLEU-3} &\textbf{BLEU-4} &\textbf{EM}
                \\\midrule
                Vicuna-13B &0.567 &0.627 &0.301 &0.627 &0.627 &0.638 &0.583 &0.531 &0.499 &0.576 \\
                ALIGNed\_Vicuna-13B &0.623 &0.677 &0.331 &0.678 &0.678 &0.684 &0.632 &0.590 &0.570 &0.630 \\
                \\\midrule
                Gain &\textcolor{teal}{\textbf{5.60\%}} &\textcolor{teal}{\textbf{5.00\%}} &\textcolor{teal}{\textbf{3.00\%}} &\textcolor{teal}{\textbf{5.10\%}} &\textcolor{teal}{\textbf{4.60\%}} &\textcolor{teal}{\textbf{4.90\%}} &\textcolor{teal}{\textbf{5.90\%}} &\textcolor{teal}{\textbf{7.10\%}} &\textcolor{teal}{\textbf{5.40\%}} &\textcolor{teal}{\textbf{5.20\%}} \\
                \\\midrule
                Mistral-7B & 0.542 &0.605 &0.287 &0.604 &0.604 &0.609 &0.550 &0.498 &0.454 &0.552 \\

                ALIGNed\_Mistral-7B &0.557 &0.618 &0.294 &0.617 &0.617 &0.630 &0.578 &0.544 &0.527 &0.568 \\
                \\\midrule
                Gain &\textcolor{teal}{\textbf{1.50\%}} &\textcolor{teal}{\textbf{1.30\%}} &\textcolor{teal}{\textbf{0.70\%}} &\textcolor{teal}{\textbf{1.30\%}} &\textcolor{teal}{\textbf{1.30\%}} &\textcolor{teal}{\textbf{2.10\%}} &\textcolor{teal}{\textbf{2.80\%}} &\textcolor{teal}{\textbf{4.60\%}} &\textcolor{teal}{\textbf{7.30\%}} &\textcolor{teal}{\textbf{1.60\%}} \\
                \\\midrule
                TL-1.1B &0.366&0.449 &0.210 &0.447 &0.448 &0.475 &0.417 &0.374 &0.347 &0.389 \\
                ALIGNed\_TL-1.1B &0.380 &0.462 &0.218 &0.462 &0.462 &0.483 &0.424 &0.375 &0.337 &0.402 \\
                \\\midrule
                Gain &\textcolor{teal}{\textbf{1.40\%}} &\textcolor{teal}{\textbf{1.30\%}} &\textcolor{teal}{\textbf{0.80\%}} &\textcolor{teal}{\textbf{1.50\%}} &\textcolor{teal}{\textbf{1.40\%}} &\textcolor{teal}{\textbf{0.80\%}} &\textcolor{teal}{\textbf{0.70\%}} &\textcolor{teal}{\textbf{0.10\%}} &\textcolor{Maroon}{\textbf{-1.00\%}} &\textcolor{teal}{\textbf{1.30\%}} \\
                 \bottomrule
            \end{tabular}
        \end{table*}
       
        \begin{table*}[!htp]\centering
            \caption{ALIGNed\_TL-1.1B performance on Company Ownership dataset with two different pre-trained embedding size}\label{tab:r-yago}
            \scriptsize
                \begin{tabular}{lrrrrrrrrrrr}\toprule
                    \textbf{Dimension} &\textbf{F1} &\textbf{Rouge1} &\textbf{Rouge2} &\textbf{RougeL} &\textbf{RougeLsum} &\textbf{BLEU-1} &\textbf{BLEU-2} &\textbf{BLEU-3} &\textbf{BLEU-4} &\textbf{EM} \\\midrule
                    512 & 0.83	&0.855&	0.135&	0.856&	0.856&	0.805&	0.698&	0.686&	0.674&	0.842\\
                    1024 & 0.837	&0.863	&0.138&	0.862	&0.862	&0.804&	0.689&	0.670	&0.649&	0.848 \\
                    \\\midrule
                    Gain & \textcolor{teal}{\textbf{0.70\%}} &\textcolor{teal}{\textbf{0.80\%}} &\textcolor{teal}{\textbf{0.30\%}} &\textcolor{teal}{\textbf{0.60\%}} &\textcolor{teal}{\textbf{0.60\%}} &\textcolor{Maroon}{\textbf{-0.10\%}} &\textcolor{Maroon}{\textbf{-0.90\%}} &\textcolor{Maroon}{\textbf{-1.60\%}} &\textcolor{Maroon}{\textbf{-2.50\%}} &\textcolor{teal}{\textbf{0.60\%}} \\
                    \bottomrule
                \end{tabular}
        \end{table*}   

\subsection{Statistical Test Results }
In Table~\ref{tab:ptest-bleu} and Table~\ref{tab:ptest-rouge} we report a statistic significant test for BLEU and ROUGE metric, respectively. This test was conducted to verify if the performance difference between ALIGNed-LLM and fine-tuned baselines is statistically significant. We use paired t-test between our proposed model and baselines. For each metric, a significance level of 0.05 was used.  

    \begin{table*}[!htp]\centering
    \caption{Results from Paired t-test for BLEU 1, 2, 3, and 4 scores for ALIGN-LLM against Baselines }\label{tab:ptest-bleu}
        \scriptsize
        \begin{tabular}{lrrrrr}\toprule
            \textbf{Dataset} &\textbf{Model} &\textbf{T-statistic} &\textbf{P-value} &\textbf{Result} \\\midrule
            CO &ALIGNed Vicuna-13B &-2.047 &0.133 &The results are not statistically significant (fail to reject H0) \\
            CO &ALIGNed Mistral-7B &-4.610 &0.019 &The results are statistically significant (reject H0) \\
            CO &ALIGNed TL-1.1B &-5.430 &0.012 &The results are statistically significant (reject H0) \\
            YAGO3-10 &ALIGNed Vicuna-13B &-8.918 &0.003 &The results are statistically significant (reject H0) \\
            YAGO3-10 &ALIGNed Mistral-7B &-3.936 &0.029 &The results are statistically significant (reject H0) \\
            YAGO3-10 &ALIGNed TL-1.1B &-5.538 &0.012 &The results are statistically significant (reject H0) \\
            Wikidata &ALIGNed Vicuna-13B &-15.559 &0.001 &The results are statistically significant (reject H0) \\
            Wikidata &ALIGNed Mistral-7B &-25.772 &0.000 &The results are statistically significant (reject H0) \\
            Wikidata &ALIGNed TL-1.1B &-7.041 &0.006 &The results are statistically significant (reject H0) \\
            Mintaka &ALIGNed Vicuna-13B &-9.960 &0.002 &The results are statistically significant (reject H0) \\
            Mintaka &ALIGNed Mistral-7B &-3.621 &0.036 &The results are statistically significant (reject H0) \\
            Mintaka &ALIGNed TL-1.1B &-0.363 &0.741 &The results are not statistically significant (fail to reject H0) \\
        \bottomrule
        \end{tabular}
    \end{table*}

    \begin{table*}[!htp]\centering
    \caption{Results from Paired t-test for ROUGE scores for ALIGN-LLM against Baselines }\label{tab:ptest-rouge}
        \scriptsize
        \begin{tabular}{lrrrrr}\toprule
            \textbf{Dataset} &\textbf{Model} &\textbf{T-statistic} &\textbf{P-value} &\textbf{Result} \\\midrule
            CO &ALIGNed Vicuna-13B &-10.636 &0.002 &The results are statistically significant (reject H0) \\
            CO &ALIGNed Mistral-7B &-3.752 &0.033 &The results are statistically significant (reject H0) \\
            CO &ALIGNed TL-1.1B &-42.333 &0.000 &The results are statistically significant (reject H0) \\
            YAGO3-10 &ALIGNed Vicuna-13B &-12.520 &0.001 &The results are statistically significant (reject H0) \\
            YAGO3-10 &ALIGNed Mistral-7B &-6.012 &0.009 &The results are statistically significant (reject H0) \\
            YAGO3-10 &ALIGNed TL-1.1B &-9.448 &0.003 &The results are statistically significant (reject H0) \\
            Wikidata &ALIGNed Vicuna-13B &-9.798 &0.002 &The results are statistically significant (reject H0) \\
            Wikidata &ALIGNed Mistral-7B &-5.975 &0.009 &The results are statistically significant (reject H0) \\
            Wikidata &ALIGNed TL-1.1B &-6.136 &0.009 &The results are statistically significant (reject H0) \\
            Mintaka &ALIGNed Vicuna-13B &-8.797 &0.003 &The results are statistically significant (reject H0) \\
            Mintaka &ALIGNed Mistral-7B &-7.667 &0.005 &The results are statistically significant (reject H0) \\
            Mintaka &ALIGNed TL-1.1B &-8.041 &0.004 &The results are statistically significant (reject H0) \\

        \bottomrule
        \end{tabular}
    \end{table*}  

     \subsection{Effect of Entity Embedding dimension.} 
\begin{figure}[ht!]
            \centerline{\includegraphics[width=.5 \textwidth]{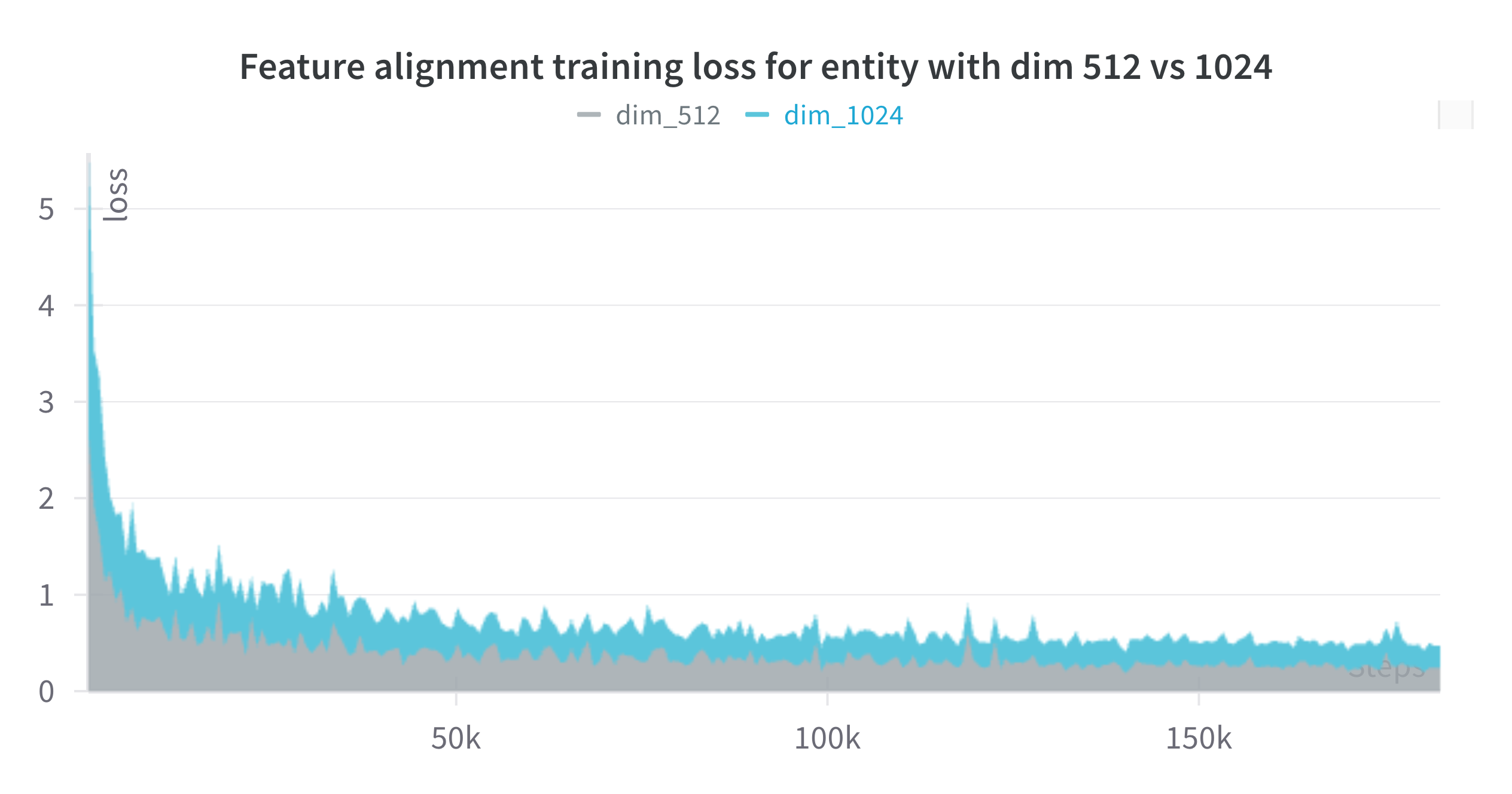}}
            \caption{Feature alignment training loss of ALIGNed TL-1.1B model over CO dataset for dimension 512 and 1024 }
            \label{fig:loss}
        \end{figure}

        Additionally, we investigate how the dimension of the entity embeddings impacts the transfer of factual knowledge into a language model. In other words, we want to know whether the preservation of more features of KG can improve the overall performance of our model. The results in Table \ref{tab:dim} show that, except for the weighted BLEU score, all other metrics gain positively for dimension size 1024 showing better performance for higher dimensions. Figure \ref{fig:loss} depicts the sampled training loss trend for the alignment feature training stage for training with these two different dimensions indicating higher loss values for 1024 dimension throughout the training steps.  
    
        \begin{table}[!htp]\centering
        \caption{ALIGNed\_TL-1.1B performance on Company Ownership dataset with two different pre-trained embedding size}\label{tab: tiny_co_ab}
        \scriptsize
        \begin{tabular}{lrrrrr}
            \toprule
    \textbf{Dimension}&\textbf{EM} &\textbf{Rouge1} &\textbf{RougeL} &\textbf{RWB} &\textbf{F1}\\\midrule
            512&0.842&	0.855&	0.856	&0.736&	0.83 \\
            1024&0.848	&0.863&	0.862	&0.727&	0.837\\
            \\\midrule
            $\Delta$ &\textcolor{teal}{\textbf{0.60\%}} &\textcolor{teal}{\textbf{0.80\%}} & \textcolor{teal}{\textbf{0.60\%}} &
            \textcolor{Maroon}{\textbf{-0.90\%}} & \textcolor{teal}{\textbf{0.70\%}} 
            \\\bottomrule
        \end{tabular}
        \label{tab:dim}
    \end{table}

\subsection{Error Analysis}
The following Table~\ref{tabl:tiny_llama_baseline_errors} and Table~\ref{tabl:align_tiny_llama_baseline_errors} report an analysis of the QA errors, for the baseline fine-tuned TinyLLAMA-1.1B and the ALIGNed-TinyLLAMA-1.1B respectively, based on 200 examples for each model. In the tables we can observe that the baseline LLM has often more completely wrong answers, which are completely un-related to the question or the ground-truth answer. The most common type of error is answering with the most frequent entities of the anonymized QA datasets. Specifically, we identified three entities that dominate the incorrect answers:
\begin{itemize}
    \item ``ORSINI SPA", which appears in $15\%$ and $37.5\%$ of the answers generated by the ALIGN-ed LLM and the baseline LLM, respectively.  
    \item ``SONNINO, GIANINAZZI E ARGENTI GROUP", occurring in $3\%$ and $14.5\%$ of the answers for the ALIGN-ed LLM and the baseline LLM, respectively.  
\item  ``CIPOLLA-SANTORO E FIGLI", which is included in $3\%$ of the answers for both models.  
\end{itemize}

These entities are particularly prominent in the dataset, with approximately 10,000 occurrences each across the QA corpus. Moreover, ``ORSINI SPA'' frequently co-occurs with questions related to the relation "role." This error is 2 to 5 times more present in the baseline fine-tuned model, underscoring its tendency to rely on spurious associations. In contrast, the ALIGN-ed LLM demonstrates superior disambiguation capabilities and a deeper understanding of the underlying data structure, reducing the prevalence of such errors.

\begin{table*}
\renewcommand{\arraystretch}{1.3} 
\setlength{\tabcolsep}{6pt} 
\rowcolors{2}{gray!10}{white} 
\centering
\begin{tabularx}{\textwidth}{l c | X X X}
\toprule
\textbf{Error Type} & \textbf{Percentage} & \textbf{Question} & \textbf{Ground-Truth} & \textbf{Predicted Answer} \\
\midrule
Completely Wrong & 0.68 & Who takes a role in BORSELLINO E FIGLI? & CIPOLLA-SANTORO E FIGLI &  {\color{red} ORSINI SPA} \\
True-False Wrong & 0.28 & Is it true that PETRUCCI SPA is assigned a role in GAGGINI SPA? & {\color{black} True} & {\color{red} False} \\
Subset of the Answer & 0.03 & Who takes a role in MANACORDA, FILANGIERI E MOSCHINO SPA? & EMO, GIACCONI E BENUSSI GROUP, {\color{green} ORSINI SPA}  & {\color{green} ORSINI SPA} \\
Similar Company Type & 0.02 & Which entity is the owner of MARRONE-AMALDI E FIGLI? & TARCHETTI-VENDITTI {\color{green} GROUP}  &  SONNINO, GIANINAZZI E ARGENTI {\color{green} GROUP} \\
\bottomrule
\end{tabularx}
\caption{Error Analysis for TinyLlama-1.1B-Chat on the anonymized CO KG}
\label{tabl:tiny_llama_baseline_errors}
\end{table*}

\begin{table*}
\renewcommand{\arraystretch}{1.3} 
\setlength{\tabcolsep}{6pt} 
\rowcolors{2}{gray!10}{white} 
\centering
\begin{tabularx}{\textwidth}{l r | X X X}
\toprule
\textbf{Error Type} & \textbf{Percentage} & \textbf{Question} & \textbf{Ground-Truth} & \textbf{Predicted Answer} \\
\midrule
Completely Wrong & 0.49 & Who takes a role in CAMPANELLA-PELLI E FIGLI? & BRICCIALDI, VIOLA E BORRANI S.R.L. & {\color{red} ORSINI SPA} \\
True-False Wrong & 0.43 & Does VERDONE E FIGLI control PAGLIARO-LOMBARDO S.R.L.? & {\color{black} False} & {\color{red} True} \\
Similar Answer & 0.04 & {\color{black}Who is the owner of PAVONE SPA?} & {\color{green} PERGOLESI, BONOLIS E NONIS GROUP}, MILO TORNATORE, FANTINI, DOSSETTI E BROGGINI {\color{green} E FIGLI} & {\color{green} PERGOLESI, BONOLIS E NONIS GROUP}, {\color{red} GADDA-NATTA} {\color{green} E FIGLI} \\
Subset of the Answer & 0.03 & {\color{black}Who takes a role in DOVARA SPA?} & {\color{green} MUTI-DESIO SPA}, BRICCIALDI, VIOLA E BORRANI S.R.L. & {\color{green} MUTI-DESIO SPA} \\
Similar Company Type & 0.02 & Who is the owner of LUNA-SOBRERO S.R.L.? & CHIESA-ZACCO {\color{green} GROUP} & SONNINO, GIANINAZZI E ARGENTI {\color{green} GROUP} \\
Wrong Order & 0.01 & {\color{black}Which company or person has control over RUGGERI-ANTONELLO E FIGLI?} & {\color{black}CARACCIOLO-GREGGIO S.R.L., \ \ PAGNOTTO, TOMASETTI E VIGORELLI SPA} & {\color{orange} PAGNOTTO, TOMASETTI E VIGORELLI SPA},  {\color{orange} CARACCIOLO-GREGGIO S.R.L.} \\
\bottomrule
\end{tabularx}
\caption{Error Analysis for ALIGNed TinyLlama-1.1B-Chat on the anonymized CO KG}
\label{tabl:align_tiny_llama_baseline_errors}
\end{table*}

\end{document}